\documentclass[11pt]{article}

\usepackage[final]{acl}

\usepackage{amsmath,amsfonts,bm}

\def\eqref#1{equation~\ref{#1}}
\def\1{\bm{1}}
\newcommand{\train}{\mathcal{D}}

\DeclareMathAlphabet{\mathsfit}{\encodingdefault}{\sfdefault}{m}{sl}
\SetMathAlphabet{\mathsfit}{bold}{\encodingdefault}{\sfdefault}{bx}{n}

\usepackage{times}
\usepackage{latexsym}

\usepackage[T1]{fontenc}
\usepackage[utf8]{inputenc}

\usepackage{microtype}

\usepackage{inconsolata}

\usepackage{algorithm}
\usepackage{algpseudocode}
\usepackage{graphicx}
\usepackage{booktabs}
\usepackage{subcaption}
\usepackage[misc]{ifsym}

\newtheorem{definition}{Definition}[section]

\title{Do Not Step Into the Same River Twice: \\ Learning to Reason from Trial and Error}

\author{
    \textbf{Chenming Tang}$^{1,2,*}$\quad
    \textbf{Hsiu-Yuan Huang}$^{1,2,3,*}$\\
    \textbf{Weijie Liu}$^{3,\dag}$\quad
    \textbf{Clive Bai}$^{3}$\quad
    \textbf{Saiyong Yang}$^{3}$\quad
    \textbf{Yunfang Wu}$^{1,2,\dag}$ \\
    $^{1}$National Key Laboratory for Multimedia Information Processing, Peking University \\ 
    $^{2}$School of Computer Science, Peking University \quad
    $^{3}$LLM Department, Tencent \\
    \small{\href{mailto:tangchenming@stu.pku.edu.cn}{tangchenming@stu.pku.edu.cn}\quad\href{mailto:jagerliu@tencent.com}{jagerliu@tencent.com}\quad\href{mailto:wuyf@pku.edu.cn}{wuyf@pku.edu.cn}}\\
    \small{$^*$Work done during an internship at Tencent \quad $^\dag$Corresponding authors}
}

\begin{document}

\maketitle

\begin{abstract}
Reinforcement learning with verifiable rewards (RLVR) has significantly boosted the reasoning capability of language models (LMs). However, existing RLVR approaches train LMs based on their own on-policy responses and are constrained by the initial capability of LMs, thus prone to exploration stagnation, in which LMs fail to solve more training problems and cannot further learn from the training data. Some approaches try to address this by leveraging off-policy solutions to training problems, but rely on external expert guidance that is limited in availability and scalability. In this work, we propose LTE (Learning to reason from Trial and Error), an approach that hints LMs with their previously self-made mistakes, not requiring any external expert guidance. Experiments validate the effectiveness of LTE, which outperforms the normal group relative policy optimization (GRPO) by 5.02 in Pass@1 and 9.96 in Pass@k on average across six mathematical reasoning benchmarks for Qwen3-8B-Base and even performs better than methods that require external guidance. Further analysis confirms that LTE successfully mitigates exploration stagnation and enhances both exploitation and exploration during training. Our code is available at \url{https://github.com/JamyDon/LTE}.
\end{abstract}

\section{Introduction}

\emph{\quad You cannot step into the same river twice.}

\rightline{---Heraclitus}

Language models (LMs) have made significant breakthroughs in reasoning capability~\cite{r1, qwen3, k2}, leveraging long chain-of-thoughts to perform test-time computing and remarkably benefiting reasoning-intensive tasks. The awesome reasoning capability of LMs is unlocked primarily by reinforcement learning with verifiable rewards (RLVR), in which the correctness of LMs' responses can be objectively and automatically verified. However, existing RLVR approaches optimize the LM based on its own on-policy rollouts and are constrained by the LM's initial capability, thus suffering from \textit{exploration stagnation}~\cite{StepHint}. To be specific, if the LM encounters a training sample sufficiently difficult to be beyond its capability upper bound, it fails to produce any correct solutions and thus receives no positive training signal from this sample. In this way, the LM can never solve problems that it could not initially (with reasonably many rollouts) solve and is bounded by itself.

\begin{figure}[tbp]
    \centering
    \includegraphics[width=1.0\linewidth]{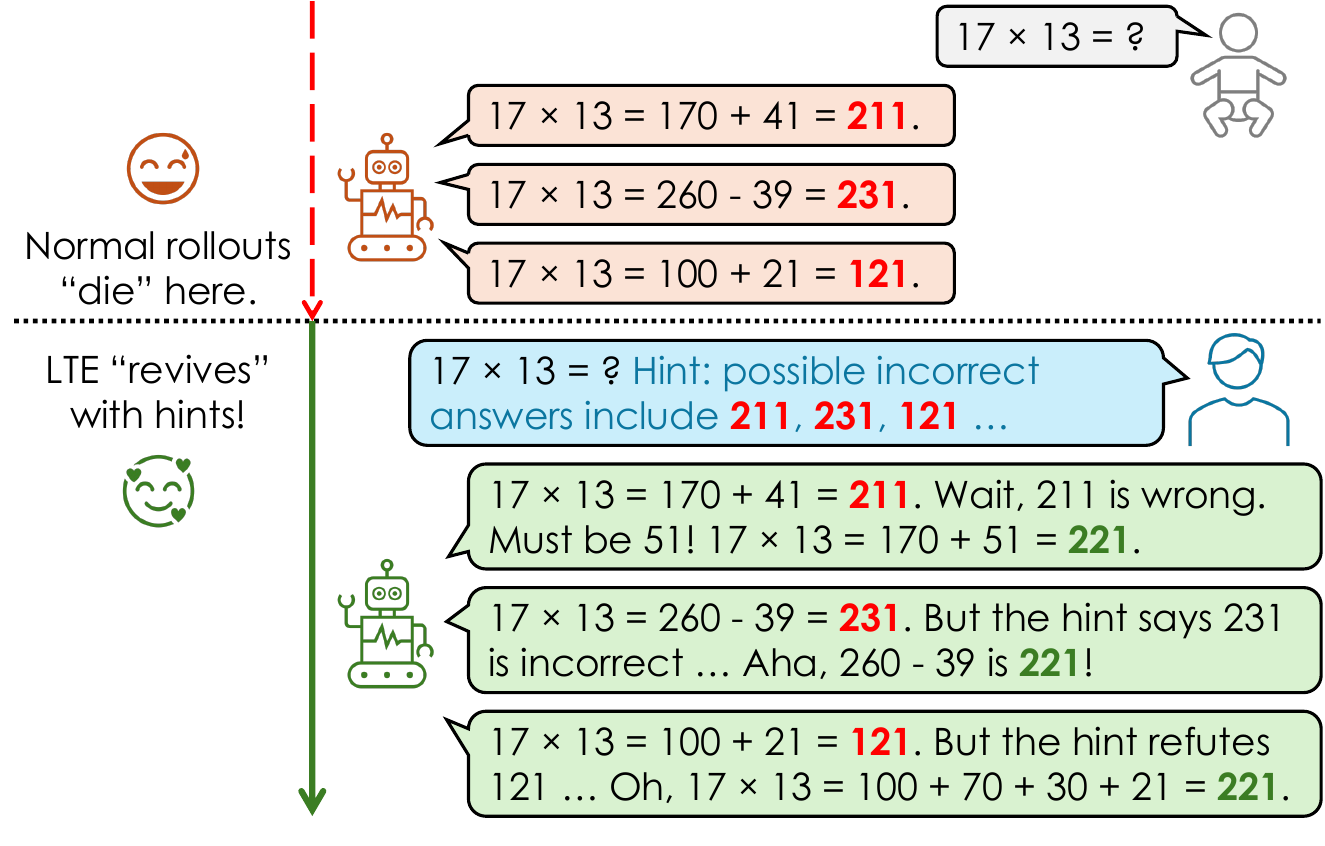}
    \caption{For difficult problems the policy model fails to solve, conventional methods simply go on with nothing gained. On the contrary, LTE tries to fix the mess with immediate hints from the model's own errors.}
    \label{fig:teaser}
\end{figure}

To address this, some introduce external guidance to help the LM break through its upper bound~\cite{luffy, ReLIFT, StepHint, chord, hpt}. However, these either utilize human-annotated ground truth solutions that suffer from high cost and limited scalability, or require correct reasoning chains generated by stronger LMs, which can be unavailable in certain cases like training flagship models. In other words, these methods are more like a palliative than a remedy for exploration stagnation.

In this work, we propose LTE (\textbf{\underline{L}}earning to reason from \textbf{\underline{T}}rial and \textbf{\underline{E}}rror), a novel method that leverages the LM's own trials as a hint, without explicit external guidance\footnote{In this paper, ``explicit external guidance'' refers to guidance of detailed reasoning traces that is less accessible than merely ground truth answers.}. For problems that no rollout passes the verification (namely \textit{none-pass} samples), we collect the incorrect answers generated from the rollouts and perform extra rollouts with these as in-context hints, warning the LM not to step into the same river twice (\textit{i.e.}, fall into errors it is prone to). If there are overly long responses that were truncated in the initial rollouts, we further prompt the LM to think concisely to reduce prolixity. In this way, LTE increases the chance of obtaining correct solutions and provides meaningful training signals for these stagnated samples (Figure~\ref{fig:teaser}). This is similar to the learning process of human. When a student is given hints about previous mistakes, he will keep away from these and be more likely to obtain the correct solution.

Our experiments validate the effectiveness of LTE, which outperforms baseline methods in both Pass@1 and Pass@k. For Qwen3-8B-Base, LTE outperforms its counterpart which simply performs extra rollouts by \textbf{+7.29} and \textbf{+10.04} in Pass@1 and Pass@k averaged across six mathematical benchmarks, respectively. The entropy-loss-enabled variant LTE$\dag$ even exhibits higher scores (\textbf{+2.41} Pass@1 and \textbf{+2.15} Pass@k) than LUFFY~\cite{luffy}, a mixed-policy method requiring external solutions from stronger LMs.

Analysis on the training data proves that LTE mitigates exploration stagnation by consistently reducing the number of unsolved samples while keeping more learnable \textit{some-pass} samples. Also, LTE achieves a higher upper bound of both Pass@1 and Pass@k during training, keeps a relatively considerable level of entropy in the long tail, and encourages exploration based on test-time deep thinking by increasing the response length. These confirm LTE implicitly elicits the internal exploration capability in LMs while also enhancing exploitation.

Our contributions are as follows:

\begin{itemize}
\item We present LTE, which addresses exploration stagnation of RLVR using LMs' own trial and error, independent of any explicit external guidance from humans or stronger LMs.
\item We empirically validate the effectiveness of LTE, which showcases the most outstanding performance among compared methods across two LMs in both Pass@1 and Pass@k.
\item We confirm that LTE successfully mitigates exploration stagnation by remarkably reducing \textit{none-pass} samples in the training dataset and elicits the internal exploration capability in LMs by keeping a relatively considerable entropy while encouraging test-time deep thinking with more tokens.
\end{itemize}

This paper is organized as follows. We introduce the preliminary of RLVR and exploration stagnation in Section~\ref{sec:preliminary} and present our proposed LTE in Section~\ref{sec:method}. Then, we list our experimental setup in Section~\ref{sec:setup} and show the experimental results in Section~\ref{sec:result}. Finally, we discuss related work in Section~\ref{sec:related-work} and conclude our study in Section~\ref{sec:conclusion}.

\begin{figure*}[htbp]
    \centering
    \includegraphics[width=1.0\linewidth]{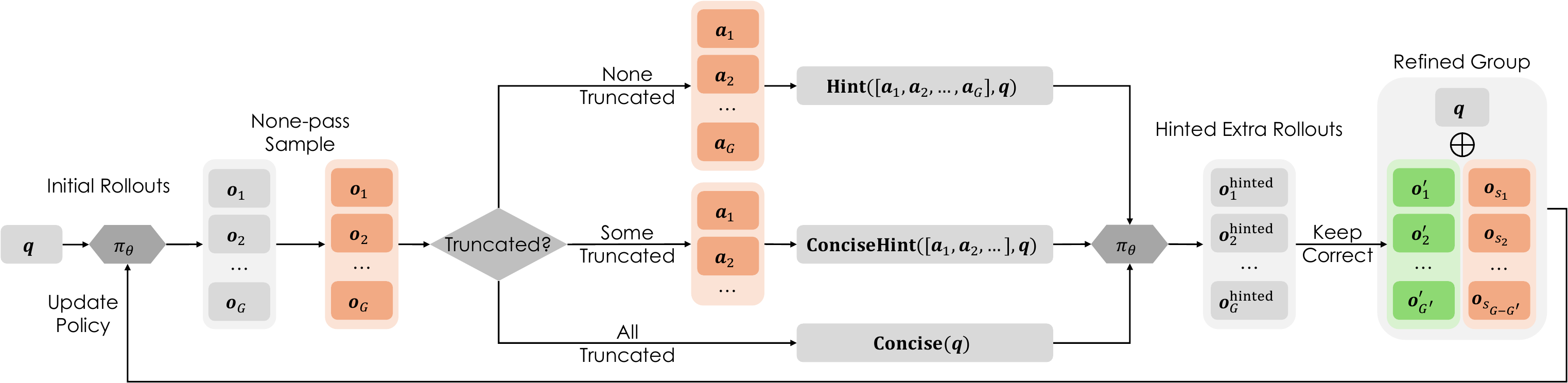}
    \caption{The framework of LTE. For \textit{none-pass} samples, we extract the LM's generated incorrect answers as hints for the extra rollouts. We do nothing to other samples which are omitted in the figure for simplicity.}
    \label{fig:method}
\end{figure*}

\section{Preliminary}\label{sec:preliminary}
\subsection{Reinforcement Learning with Verifiable Rewards (RLVR)}
RLVR is an emergent reinforcement paradigm for the post-training of LMs, aiming to enhance the capability of LMs, especially in reasoning. In the RLVR framework, an LM $\pi_\theta$ is trained on tasks where the correctness of an outcome can be automatically verified. Group relative policy optimization (GRPO)~\cite{grpo} is the \textit{de facto} standard algorithm for RLVR due to its effectiveness and simplicity. Given an input query $q\sim\train$, $G$ rollouts $\{o_1, o_2, ..., o_G\}$ are sampled from $\pi_\theta(\cdot|q)$ and rewarded by a reward function $R$. Then GRPO, adopting the token-level loss proposed by~\citet{dapo}, maximizes the following objective:

\small
\begin{equation}
\begin{aligned}
\label{eq:grpo}
\mathcal{J}_{\text{GRPO}}(\theta)=
&\mathbb{E}_{q,\{o_i\}}\Bigl[
 \frac{1}{Z}\!\sum_{i=1}^{G}\!\sum_{t=1}^{|o_i|}
 \Bigl(\operatorname{CLIP}(r_{i,t}(\theta), \hat{A}_{i,t}, \varepsilon)\;
  \\
 &\ \ \ \ \ \ \ \ \ \ \ \ \ \ \ \ \ \ \ \ \ \ \ \ \ -\beta\,\mathrm{D}_{\mathrm{KL}}(\pi_{\theta}\Vert\pi_{\text{ref}})\Bigr)
\Bigr],
\end{aligned}
\end{equation}
\normalsize
where $Z$ is the normalization factor:
\small
\begin{equation}\label{eq:normalization}
Z=\sum_{i=1}^{G}|o_i|,
\end{equation}
\normalsize
$r_{i,t}(\theta)$ is the importance sampling ratio:
\small
\begin{equation}\label{eq:ratio}
    r_{i,t}(\theta)
= \frac{\pi_{\theta}\bigl(o_{i,t}\mid q,\, o_{i,<t}\bigr)}
       {\pi_{\theta_{\mathrm{old}}}\bigl(o_{i,t}\mid q,\,o_{i,<t}\bigr)},
\end{equation}
\normalsize
\(\hat{A}_{i,t}\) is the group-relative advantage:
\small
\begin{equation}\label{eq:adv}
\hat{A}_{i,t}
= \frac{R(o_i) \;-\;\mathrm{Mean}\bigl(\{R(o_j)\}_{j=1}^G\bigr)}
       {\mathrm{Std}\bigl(\{R(o_j)\}_{j=1}^G\bigr)},
\end{equation}
\normalsize
and $\mathrm{D}_{\mathrm{KL}}(\pi_{\theta}\Vert\pi_{\text{ref}})$ is the KL penalty term.

\subsection{Exploration Stagnation}
A major limitation of existing RLVR approaches is that all the rollouts are on-policy and the LM is constrained by its initial capability, failing to break through its own upper bound. This is known as \textit{exploration stagnation}~\cite{StepHint}.

To be specific, in the typical $0/1$ reward setting, if a training problem $q$ is too difficult for the LM to answer (\textit{i.e.}, beyond its capability upper bound), each of the G rollouts receives a zero reward. Then all the group-relative advantages degenerate to zero, which means that the policy learns nothing from such a difficult problem.

Formally, after ignoring the clip function and KL term, the GRPO gradient can be represented as:

\small
\begin{equation}
\begin{aligned}
& \nabla_{\theta} \mathcal{J}_{\text{GRPO}}(\theta)=\mathbb{E}_{q, \{o_i\}}\\
& \left[ \sum_{i=1}^{G} \left( \sum_{t=1}^{|o_i|} \nabla_{\theta} \log \pi_{\theta}(o_{i,t}\mid q,\, o_{i,<t}) \cdot \hat{A}_{i,t} \right) \right].
\end{aligned}
\end{equation}
\normalsize

In \textit{none-pass} samples, we have $R(o_i) = 0$ for all $i$. Thus, we get $\hat{A}_{i,t} = 0$ for all $i$ and $t$. In this way, $\nabla_{\theta} \mathcal{J}_{\text{GRPO}} = 0$, which means the policy learns nothing from such samples.

\section{LTE: Learning from Trial and Error}\label{sec:method}
The framework of LTE is shown in Figure~\ref{fig:method} and its training workflow is summarized in Algorithm~\ref{alg:lte}. We provide a theoretical analysis in Appendix~\ref{app:theory} demonstrating that LTE increases the chance of reaching the correct solutions.

\subsection{Hinted Extra Rollouts}\label{sec:hinted-extra-rollouts}

To address exploration stagnation, a trivial approach is to perform extra rollouts~\cite{brorl}. For example, when all of the $G$ rollouts fail for the problem $q$, another $G$ rollouts $\{o_{G+1}, o_{G+2}, ..., o_{2G}\}$ are sampled to increase the chance of obtaining correct rollouts.

\begin{figure}[tbp]
\centering
    \begin{subfigure}[t]{0.48\textwidth}
        \centering
        \includegraphics[width=\linewidth]{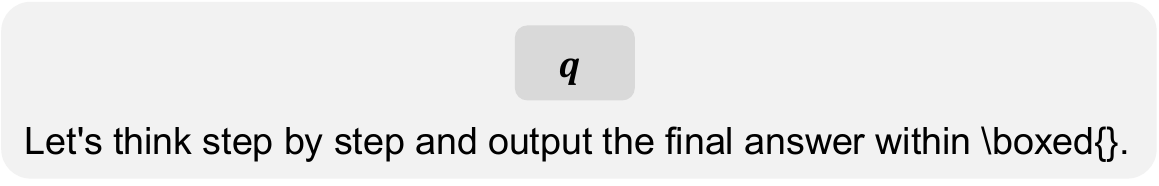}
        \caption{The normal prompt template, namely $\textbf{Prompt}(\cdot)$.}
        \label{subfig:prompt-vanilla}
    \end{subfigure}

    \begin{subfigure}[t]{0.48\textwidth}
        \centering
        \includegraphics[width=\linewidth]{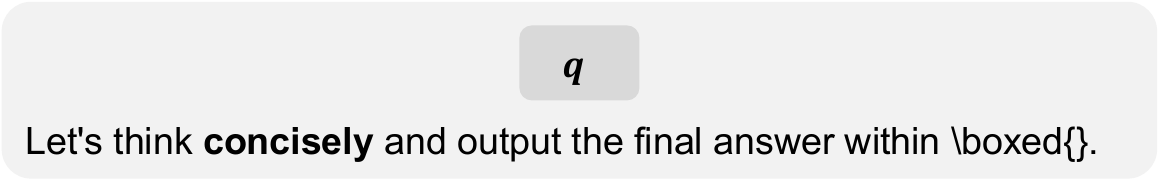}
        \caption{The prompt template for \textit{all-truncated} \textit{none-pass} samples, namely $\textbf{Concise}(\cdot)$.}
        \label{subfig:prompt-concise}
    \end{subfigure}

    \begin{subfigure}[t]{0.48\textwidth}
        \centering
        \includegraphics[width=\linewidth]{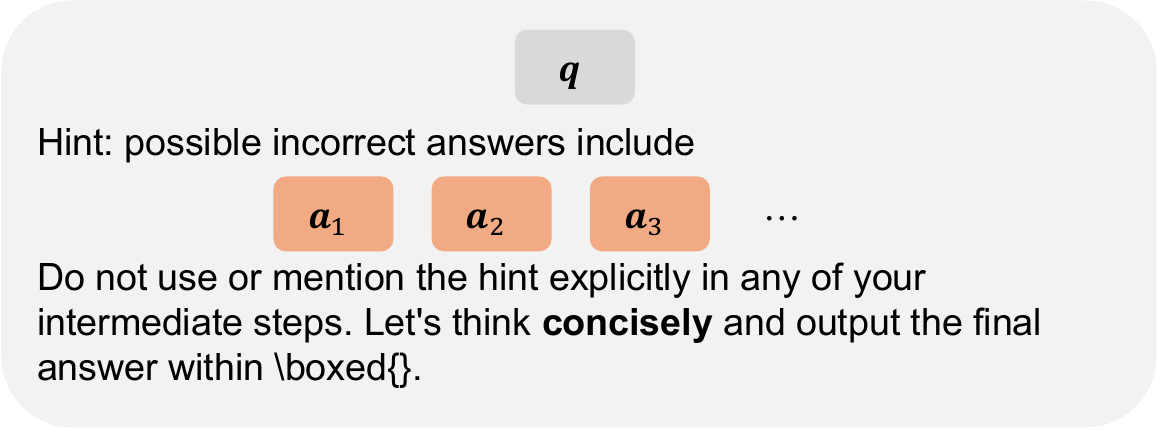}
        \caption{The prompt template for \textit{some-truncated} \textit{none-pass} samples, namely $\textbf{ConciseHint}(\cdot)$.}
        \label{subfig:prompt-concise-hint}
    \end{subfigure}

    \begin{subfigure}[t]{0.48\textwidth}
        \centering
        \includegraphics[width=\linewidth]{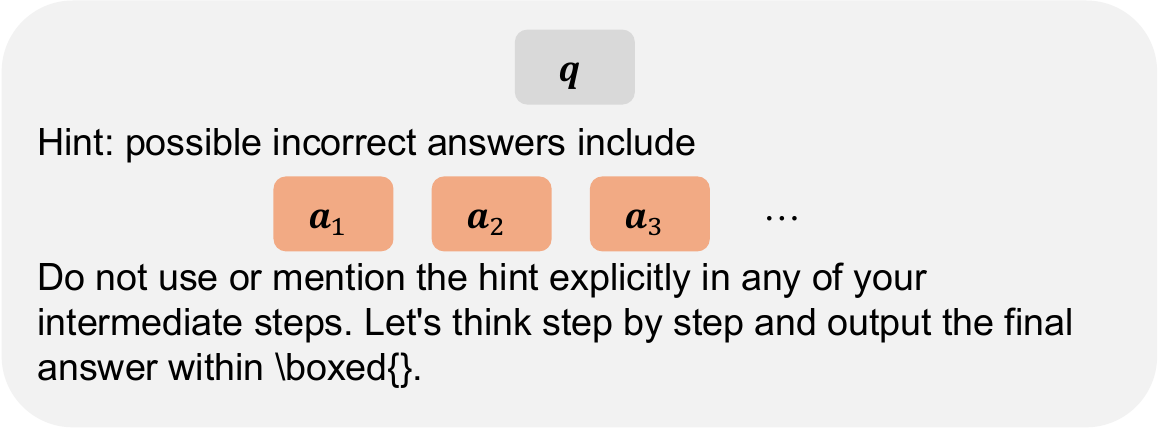}
        \caption{The prompt template for \textit{none-truncated} \textit{none-pass} samples, namely $\textbf{Hint}(\cdot)$.}
        \label{subfig:prompt-hint}
    \end{subfigure}
    
    \caption{The normal and hinted prompt templates.}
    \label{fig:prompt}
\end{figure}

Although the approach of vanilla extra rollouts is sensible, it makes use of no information from existing rollouts and can be inefficient as can be seen from the plain prompt template in Figure~\ref{subfig:prompt-vanilla}. Instead, we propose hinted extra rollouts with guidance from existing self-generated trials, making full use of the existing rollouts.

Concretely, when all the $G$ rollouts fail, we decide the type of hint based on the number of truncated overlong responses in the $G$ outputs.

If the problem is \textit{all-truncated}, in which all the responses are truncated due to length exceeding, we attribute the failure to the prolixity of responses and hint the LM to think concisely. The prompt template employed in the extra rollouts is demonstrated in Figure~\ref{subfig:prompt-concise}.\footnote{We treat all truncated responses as they do not contain any valid answers for the completeness of responses and consideration of implementation simplicity, albeit there may still exist extractable answers in truncated responses.}

If it is \textit{some-truncated} or \textit{none-truncated}, in which there exist responses that are not overlong, we collect the extracted answers $\{a_1, a_2, ...\}$ in these, which are all valuable possible incorrect answers that the LM is prone to. Then, we include these answers in the prompt as a hint to prevent the LM from falling into the same incorrect answers again (\textit{i.e.}, step into the same river twice). This reduces the size of the solution space for the LM, making it easier to reach the correct solution. For \textit{some-truncated} problems, we also hint the LM to think concisely to reduce prolixity. The prompt templates for \textit{some-truncated} and \textit{none-truncated} problems are demonstrated in Figure~\ref{subfig:prompt-concise-hint} and~\ref{subfig:prompt-hint}, respectively, where the model is instructed not to explicitly use or mention the hint so that the reasoning chain is kept as clean as possible.

Based on the chosen prompt template, we sample $G$ extra rollouts $\{o^{\text{hinted}}_{1}, o^{\text{hinted}}_{2}, ..., o^{\text{hinted}}_{G}\}$. 

\subsection{Mixed-policy Optimization}\label{subsec:mix-policy}

After the hints, if there exist $G'$ correct solutions $\{o'_1, o'_2, ..., o'_{G'}\}$ in the extra rollouts, we randomly replace $G'$ responses in the initial failed rollouts with these correct ones when updating the policy, ensuring the policy receives training signal from the initially insurmountable problem. The final responses for policy updating are denoted by $\{o'_1, o'_2, ..., o'_{G'}, o_{s_1}, o_{s_2}, ..., o_{s_{G-G'}}\}$, where $s_i$ denotes the index of the $i$-th remaining initial rollout after the random replacement. We discuss the necessity of this design of random replacement in Appendix~\ref{app:random-replacement}.

Note that since the correct solutions for these \textit{none-pass} samples are generated conditioned on the hinted prompt (denoted by $(\mathcal{H}_q, q)$ where $\mathcal{H}_q$ refers to the hint information for $q$) while the policy we need to optimize should generate solutions conditioning on the normal prompt only, they should be treated in an off-policy manner. We define a hinted policy as $\hat{\pi}(\cdot\mid\cdot)=:\pi_\text{old}(\cdot\mid\mathcal{H}_q, \cdot)$, which treat the hint as part of the defined policy. Then, the hinted sampling comes from the policy $\hat{\pi}$, which is equivalent to the off-policy distribution $\pi_\phi$ used by~\citet{luffy}. To calibrate gradient estimates~\cite{trpo}, the divisor of the importance sampling ratio should be based on the sampling distribution $\hat{\pi}$, which is exactly $\pi_\text{old}(\cdot\mid\mathcal{H}_q, \cdot)$. Therefore, the importance sampling for these samples is given by:

\small
\begin{equation}\label{eq:off-ratio}
        \hat{r}'_{i,t}(\theta)
= \frac{\pi_{\theta}\bigl(o'_{i,t}\mid q,\, o_{i,<t}\bigr)}
    {\pi_{\theta_{\mathrm{old}}}\bigl(o'_{i,t}\mid \mathcal{H}_q, q,\, o_{i,<t}\bigr)}.
\end{equation}
\normalsize

Following~\citet{luffy}, we adopt regularized importance sampling for policy shaping on these samples:

\small
\begin{equation}\label{eq:shaping}
        f(\hat{r}'_{i,t}(\theta))
= \frac{\hat{r}'_{i,t}(\theta)}
       {\hat{r}'_{i,t}(\theta) + \gamma},
\end{equation}
\normalsize
and perform mixed-policy GRPO:
\small
\begin{equation}
\begin{aligned}
\label{eq:mixed-policy}
\mathcal{J}_{\text{Mixed}}(\theta)=
&\mathbb{E}_{q,\{o'_i, o_{s_i}\}}\Bigl[
 \frac{1}{Z'}\!\sum_{i=1}^{G'}\!\sum_{t=1}^{|o'_i|}
 (f(\hat{r}'_{i,t}(\theta))\cdot\hat{A}'_{i,t})\;
  \\
 +\frac{1}{Z}&\!\sum_{i=1}^{G-G'}\!\sum_{t=1}^{|o_{s_i}|}
 \Bigl(\operatorname{CLIP}(r_{s_i,t}(\theta), \hat{A}_{s_i,t}, \varepsilon\Bigr)
\Bigr],
\end{aligned}
\end{equation}
\normalsize
where the normalization factors $Z'$ and $Z$ follow the definition in Equation~\ref{eq:normalization}, and the KL term is omitted in the formula for simplicity.

\begin{algorithm}[htbp]
\scriptsize
\caption{\textbf{LTE}: Learning from Trial and Error}
\label{alg:lte}
\begin{algorithmic}[1]
\Require Policy LM $\pi_\theta$, number of rollouts $G$, batch size $n$, number of training steps $T$, training data $\mathcal{D}$
\Ensure Updated policy LM $\pi_\theta$

\For{$t=1$ \textbf{to} $T$}
  \State $\mathcal{Q} \gets \{q_i \sim \mathcal{D}\}_{i=1}^{n}$ \Comment{training batch}
  \State $\mathcal{O}\gets\emptyset; \hat{A}\gets\emptyset$ \Comment{outputs and advantages}
  \For{$q\in\mathcal{Q}$}
    \State $\mathcal{O}_q \gets \{o_i\sim\pi_\theta(\cdot\mid q)\}_{i=1}^{G}$ \Comment{response sampling}
    \State $\mathcal{A}_q \gets \{\textsc{Extract}(o_i)\}_{i=1}^{G}$ \Comment{answer extraction}
    \State $\mathcal{R}_q \gets \{\textsc{Eval}(\mathcal{A}_q^{(i)},q)\}_{i=1}^{G}$ \Comment{evaluation}
    \If{$\forall r\in\mathcal{R}_q, r=0$} \Comment{\textit{none-pass} sample}
      \If{$\forall o\in\mathcal{O}_q, \textsc{Truncated}(o)=1$}
        \State $q'\gets\textsc{Concise}(q)$ \Comment{all-truncated}
      \ElsIf{$\forall o\in\mathcal{O}_q, \textsc{Truncated}(o)=0$}
        \State $q'\gets\textsc{ConciseHint}(q, \mathcal{A}_q)$ \Comment{some-truncated}
      \Else
        \State $q'\gets\textsc{Hint}(q, \mathcal{A}_q)$ \Comment{none-truncated}
      \EndIf
      \State $\mathcal{O}^\text{hinted}_q \gets \{o_i\sim\pi_\theta(\cdot\mid q')\}_{i=1}^{G}$ \Comment{hinted rollouts}
      \State $\mathcal{A}^\text{hinted}_q \gets \{\textsc{Extract}(o_i)\}_{i=1}^{G}$ \Comment{answer extraction}
    \State $\mathcal{R}^\text{hinted}_q \gets \{\textsc{Eval}(\mathcal{A}_q^{(i)},q)\}_{i=1}^{G}$ \Comment{evaluation}
    \State $\mathcal{O}^*_q \gets \{o_i\mid\mathcal{R}^{(i)}_q = 1\}_{i=1}^{G}$ \Comment{correct outputs}
    \State $\mathcal{R}^*_q \gets \{\mathcal{R}^{(i)}_q\mid\mathcal{R}^{(i)}_q = 1\}_{i=1}^{G}$ \Comment{corresponding rewards}
    \State $\mathcal{O}_q, \mathcal{R}_q\gets\textsc{Replace}(\mathcal{O}_q, \mathcal{R}_q, \mathcal{O}^*_q, \mathcal{R}^*_q)$ \Comment{replacement}
    \EndIf
    \State $\mathcal{O}\gets\mathcal{O}\cup\{\mathcal{O}_q\}$ \Comment{append current outputs}
    \State $\hat{A}\gets\hat{A}\cup\{\textsc{Advantage}(\mathcal{R}_q)\}$ \Comment{append current advantages}
  \EndFor
  \State $\pi_\theta \gets \textsc{MixedPolicyUpdate}(\pi_\theta, \mathcal{Q}, \mathcal{O}, \hat{A})$ \Comment{update policy}
\EndFor
\State \Return $\pi_\theta$

\end{algorithmic}
\end{algorithm}

\section{Experimental Setup}\label{sec:setup}
\subsection{Language Models}
We experiment with Qwen3-4B-Base and Qwen3-8B-Base~\cite{qwen3}. We do not use the Qwen2.5 herd of models~\cite{qwen25} because they may undergo a benchmark contamination issue~\cite{reasoning-memorization}. We do not include the Llama herd of models~\cite{llama3} because they show significant instability in reinforcement training and even exhibit a degenerated performance after training in our early experiments.

\subsection{Training}
We focus on mathematical reasoning and adopt OpenR1-Math-46k-8192~\cite{luffy} as our training dataset, which contains 45,792 instances. We adopt \texttt{verl}~\cite{verl} as the training framework. We sample 8 rollouts per prompt and set the temperature as 1.0. The maximum response length is 8,192 following~\citet{luffy}. We conduct training for 500 steps, with a batch size of 128. More details are  in Appendix~\ref{app:detail}.

\subsection{Evaluation}
We focus on six widely used mathematics benchmarks including MATH-500~\cite{MATH}, Minerva~\cite{Minerva}, OlympiadBench~\cite{OlympiadBench}, AMC'23, AIME'24 and AIME'25. For MATH-500, Minerva and OlympiadBench, we sample 4 times and report Mean@4 (for Pass@1) and Pass@4. For AMC'23, AIME'24 and AIME'25, we sample 16 times and report Mean@16 (for Pass@1) and Pass@16.

Note that the hints of LTE are only applied in the training stage. In evaluation, we strictly follow the standard protocol. The context length is set as 32,768 (\textit{i.e.}, the default maximum context length of the Qwen3 series of models) for evaluation to minimize the interference caused by truncated overlong responses. We set temperature as 0.6, top-p as 0.95 and top-k as 20 following~\citet{qwen3}.

% For both training and evaluation, we verify the correctness of answers with \texttt{Math-Verify}\footnote{\url{https://github.com/huggingface/Math-Verify}}.

\begin{table*}[htbp]
\small
  \centering
    \begin{tabular}{lcccccccc}
    \toprule
    \textbf{Method} & \textbf{EC} & \textbf{MATH-500} & \textbf{Minerva} & \textbf{Olympiad} & \textbf{AMC'23} & \textbf{AIME'24} & \textbf{AIME'25} & \textbf{Avg.} \\
    \midrule
    \textbf{\textsc{Qwen3-4B-Base}} & -     & 45.40 & 19.49 & 22.81 & 35.31 & 8.75  & 3.75  & 22.59 \\
    \midrule
    \; GRPO  & -     & 70.45 & \textbf{34.28} & 35.74 & 47.03 & 11.88 & 9.17  & 34.76 \\
    \; GRPO$_\text{Extra}$ & -     & 71.00 & 31.43 & 35.11 & 54.22 & 13.33 & 13.75 & 36.47 \\
    \; EvoCoT & -     & 67.05 & 30.06 & 31.33 & 49.84 & 9.79  & 8.75  & 32.80 \\
    \; LTE   & -     & \textbf{74.55} & 34.19 & \textbf{39.78} & \textbf{56.72} & \textbf{18.33} & \textbf{14.17} & \textbf{39.62} \\
    \midrule
    \; ReLIFT & EL    & 51.20 & 27.30 & 22.11 & 35.31 & 7.29  & 5.00  & 24.70 \\
    \; DAPO  & CH    & 76.10 & 35.29 & 42.96 & 67.34 & 22.08 & \textbf{22.92} & 44.45 \\
    \; LUFFY & EL    & 76.85 & 32.26 & 40.89 & 64.53 & \textbf{25.62} & 18.33 & 43.08 \\
    \; LTE$\dag$  & EL    & \textbf{77.25} & \textbf{37.04} & \textbf{44.15} & \textbf{69.38} & \textbf{25.62} & 22.50 & \textbf{45.99} \\
    \midrule
    \midrule
    \textbf{\textsc{Qwen3-8B-Base}} & -     & 60.85 & 26.10 & 27.41 & 47.34 & 11.46 & 9.17  & 30.39 \\
    \midrule
    \; GRPO  & -     & 77.70 & 37.32 & 42.07 & 68.91 & 22.08 & 15.83 & 43.99 \\
    \; GRPO$_\text{Extra}$ & -     & 75.75 & 36.95 & 40.07 & 60.47 & 22.29 & 14.79 & 41.72 \\
    \; EvoCoT & -     & 70.45 & 33.09 & 36.37 & 56.41 & 16.46 & 14.79 & 37.93 \\
    \; LTE   & -     & \textbf{78.85} & \textbf{37.87} & \textbf{45.52} & \textbf{73.91} & \textbf{30.62} & \textbf{27.29} & \textbf{49.01} \\
    \midrule
    \; ReLIFT & EL    & 68.15 & 33.73 & 32.89 & 55.62 & 15.21 & 10.21 & 35.97 \\
    \; DAPO  & CH    & 76.85 & \textbf{38.97} & 41.81 & 70.00 & 23.54 & 21.25 & 45.40 \\
    \; LUFFY & EL    & \textbf{80.20} & 36.76 & 44.74 & 70.78 & 29.17 & 21.25 & 47.15 \\
    \; LTE$\dag$  & EL    & 79.80 & 38.79 & \textbf{47.59} & \textbf{74.53} & \textbf{30.83} & \textbf{25.83} & \textbf{49.56} \\
    \bottomrule
    \end{tabular}%
  \caption{Pass@1 results, with best results \textbf{bolded}. ``EC'', ``EL'' and ``CH'' refers to entropy control, entropy loss and clip-higher, respectively. ``-'' means the method does not explicitly control the policy entropy.}
  \label{tab:pass-1}%
\end{table*}%

\subsection{Baselines and Methods}
Our evaluated baselines and methods include:
(1) \textbf{Base Model}: the performance of the base LM;
(2) \textbf{GRPO}: the normal baseline of GRPO;
(3) \textbf{GRPO$_\text{Extra}$}: simply performing vanilla extra rollouts for \textit{none-pass} samples without hints.
(4) \textbf{EvoCoT}~\cite{EvoCoT}: adopting the ground truth answer as a hint to overcome the exploration bottleneck in reinforcement learning.
(5) \textbf{ReLIFT}~\cite{ReLIFT}: alternating between reinforcement learning and supervised finetuning to inject off-policy reasoning traces for difficult problems.
(6) \textbf{DAPO}~\cite{dapo}: an advanced policy optimization algorithm with several techniques like clip-higher, dynamic sampling and overlong reward shaping.
(7) \textbf{LUFFY}~\cite{luffy}: mixed-policy optimization with the guidance from reasoning traces generated by stronger models.
(8) \textbf{LTE}: our proposed method.
(9) \textbf{LTE$\dag$}: our proposed LTE with entropy loss enabled, aiming to ensure a fairer comparison with methods involving entropy control (\textit{e.g.}, ReLIFT, DAPO and LUFFY) and study how LTE performs with better exploration promoted by entropy control.

\section{Results and Analysis}\label{sec:result}

\subsection{Main Results }
Table~\ref{tab:pass-1} presents the Pass@1 results. When trained without an entropy loss, LTE outperforms GRPO and GRPO$_\text{Extra}$ with \textbf{+4.94} and \textbf{+5.22} average improvements on the two LMs across the six benchmarks. This confirms that with the hint based on the LM's own trial and error, LTE brings effective guidance in the extra rollouts and successfully yields a superior final policy after the training process. On the contrary, merely performing vanilla extra rollouts, which makes no use of existing experience of trial and error, only brings marginal (even negative) improvement to GRPO, demonstrating the limitations of simple rollout scaling.

With the help of entropy loss, which brings better exploration, the power of LTE$\dag$ is further amplified, which exhibits the highest average scores across the six benchmarks among all the compared methods. It is noteworthy that LTE$\dag$, without any explicit external guidance, even performs better than LUFFY, which requires reasoning traces from stronger models, with an average gain of \textbf{+2.66} points.

Table~\ref{tab:pass-k} presents the Pass@k results which focus on the exploration upper bound (to a certain degree). Compared with the base model, GRPO and GRPO$_\text{Extra}$ improve Pass@k relatively marginally. For instance, the latter brings improvements of \textbf{+2.39} and \textbf{+3.87} for the two base models, respectively. On the contrary, LTE significantly enhances exploration with \textbf{+7.30} and \textbf{+13.91} higher Pass@k scores for the two LMs, which are \textbf{3.1×} and \textbf{3.6×} of the improvements provided by GRPO$_\text{Extra}$. Furthermore, LTE$\dag$ unlocks a higher exploration upper bound, with \textbf{+13.23} and \textbf{+14.71} improvements over the two base models, outperforming all the compared baselines including DAPO (by \textbf{+4.40} on average) and LUFFY (by \textbf{+1.79} on average) which either explicitly encourage exploration or utilize guidance of external gold reasoning traces.

Note that we run the exact training scripts provided by~\citet{ReLIFT} and~\citet{EvoCoT} but get relatively poor results for ReLIFT and EvoCoT. We attribute these to the issue of compatibility of foundation models and leave the investigation of the underlying causes for future work.

Combining the Pass@1 and Pass@k results, we conclude that, with the guidance from the LM's own behavior only, LTE encourages both exploitation and exploration during the RLVR process and unlocks higher performance without any explicit external expert guidance.

\begin{table*}[htbp]
\small
  \centering
    \begin{tabular}{lcccccccc}
    \toprule
    \textbf{Method} & \textbf{EC} & \textbf{MATH-500} & \textbf{Minerva} & \textbf{Olympiad} & \textbf{AMC'23} & \textbf{AIME'24} & \textbf{AIME'25} & \textbf{Avg.} \\
    \midrule
    \textbf{\textsc{Qwen3-4B-Base}} & -     & 69.80 & 37.87 & 39.70 & 82.50 & 33.33 & 26.67 & 48.31 \\
    \midrule
    \; GRPO  & -     & 76.40 & 40.81 & 43.85 & \textbf{85.00} & \textbf{33.33} & 23.33 & 50.45 \\
    \; GRPO$_\text{Extra}$ & -     & 78.20 & 40.81 & 42.67 & 82.50 & 26.67 & 33.33 & 50.70 \\
    \; EvoCoT & -     & 77.60 & 39.34 & 40.44 & 82.50 & 26.67 & 30.00 & 49.43 \\
    \; LTE   & -     & \textbf{80.60} & \textbf{41.91} & \textbf{49.48} & \textbf{85.00} & 30.00 & \textbf{46.67} & \textbf{55.61} \\
    \midrule
    \; ReLIFT & EL    & 76.60 & 42.28 & 39.26 & 85.00 & 30.00 & 26.67 & 49.97 \\
    \; DAPO  & CH    & 82.40 & 44.85 & 52.00 & 92.50 & 46.67 & 40.00 & 59.74 \\
    \; LUFFY & EL    & \textbf{83.20} & 42.28 & 50.22 & \textbf{95.00} & \textbf{50.00} & 40.00 & 60.12 \\
    \; LTE$\dag$  & EL    & 82.40 & \textbf{46.69} & \textbf{53.48} & 90.00 & \textbf{50.00} & \textbf{46.67} & \textbf{61.54} \\
    \midrule
    \midrule
    \textbf{\textsc{Qwen3-8B-Base}} & -     & 77.80 & 43.38 & 43.56 & 82.50 & 40.00 & 30.00 & 52.87 \\
    \midrule
    \; GRPO  & -     & 82.40 & 44.49 & 50.67 & 90.00 & 43.33 & 30.00 & 56.82 \\
    \; GRPO$_\text{Extra}$ & -     & 82.00 & 43.75 & 48.00 & 90.00 & 46.67 & 30.00 & 56.74 \\
    \; EvoCoT & -     & 76.40 & 40.81 & 46.22 & 87.50 & 43.33 & 23.33 & 52.93 \\
    \; LTE   & -     & \textbf{85.20} & \textbf{47.06} & \textbf{54.22} & \textbf{97.50} & \textbf{70.00} & \textbf{46.67} & \textbf{66.78} \\
    \midrule
    \; ReLIFT & EL    & 79.40 & 47.43 & 45.63 & 87.50 & 50.00 & 33.33 & 57.22 \\
    \; DAPO  & CH    & 82.80 & \textbf{47.79} & 50.37 & 92.50 & 53.33 & 36.67 & 60.58 \\
    \; LUFFY & EL    & \textbf{85.60} & 47.43 & 54.52 & 95.00 & 60.00 & 50.00 & 65.43 \\
    \; LTE$\dag$  & EL    & 84.60 & \textbf{47.79} & \textbf{55.56} & \textbf{97.50} & \textbf{66.67} & \textbf{53.33} & \textbf{67.58} \\
    \bottomrule
    \end{tabular}%
  \caption{Pass@k results, with best results \textbf{bolded}. ``EC'', ``EL'' and ``CH'' refers to entropy control, entropy loss and clip-higher, respectively. ``-'' means the method does not explicitly control the policy entropy.}
  \label{tab:pass-k}%
\end{table*}%

%\subsection{Performance on the Training Data}
\subsection{Analysis on Different Training Samples}
\begin{figure}[htbp]
\centering
    \begin{subfigure}[t]{0.237\textwidth}
        \centering
        \includegraphics[width=\linewidth]{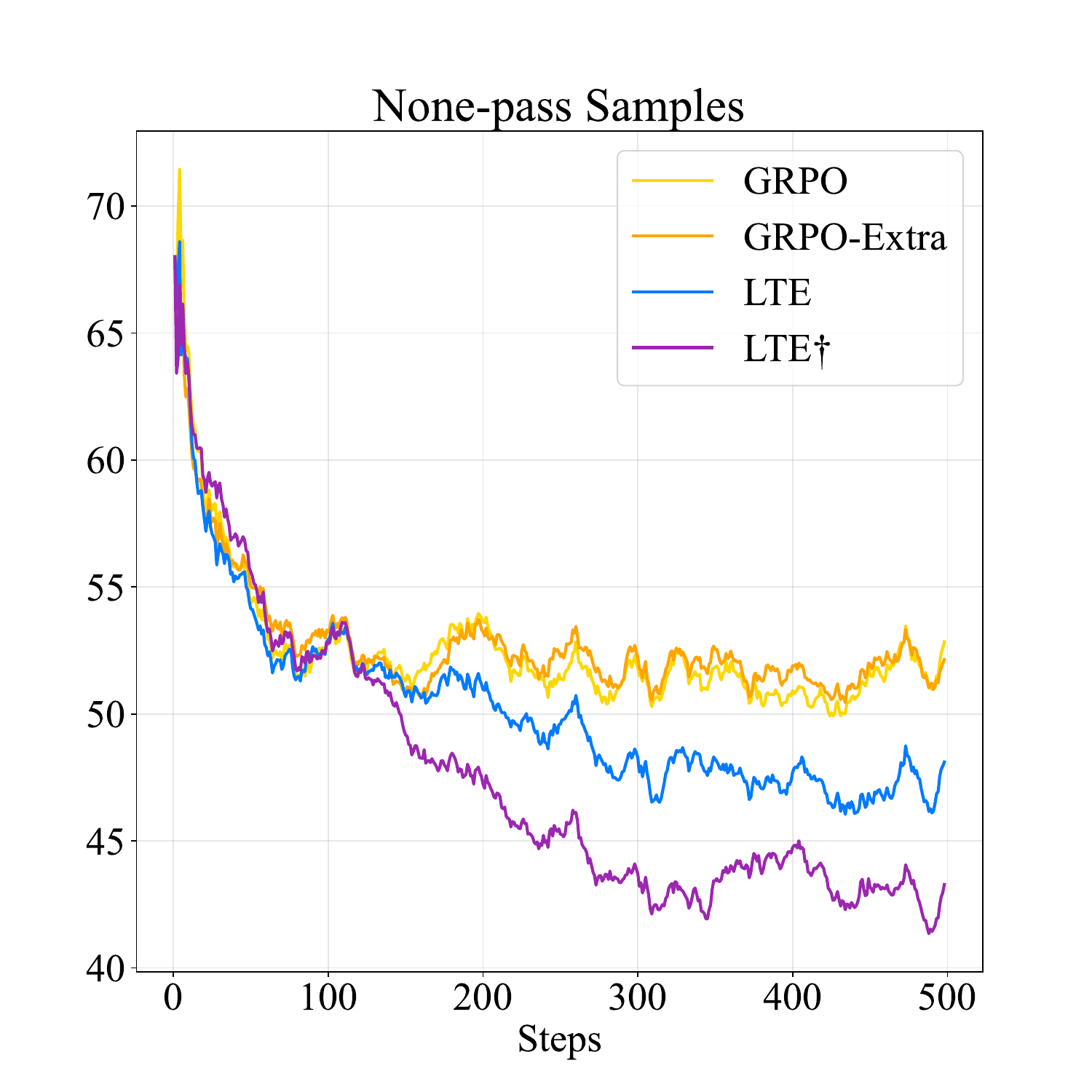}
        \caption{The number of \textit{none-pass} samples in the initial rollouts.}
        \label{subfig:none-pass}
    \end{subfigure}
    \begin{subfigure}[t]{0.237\textwidth}
        \centering
        \includegraphics[width=\linewidth]{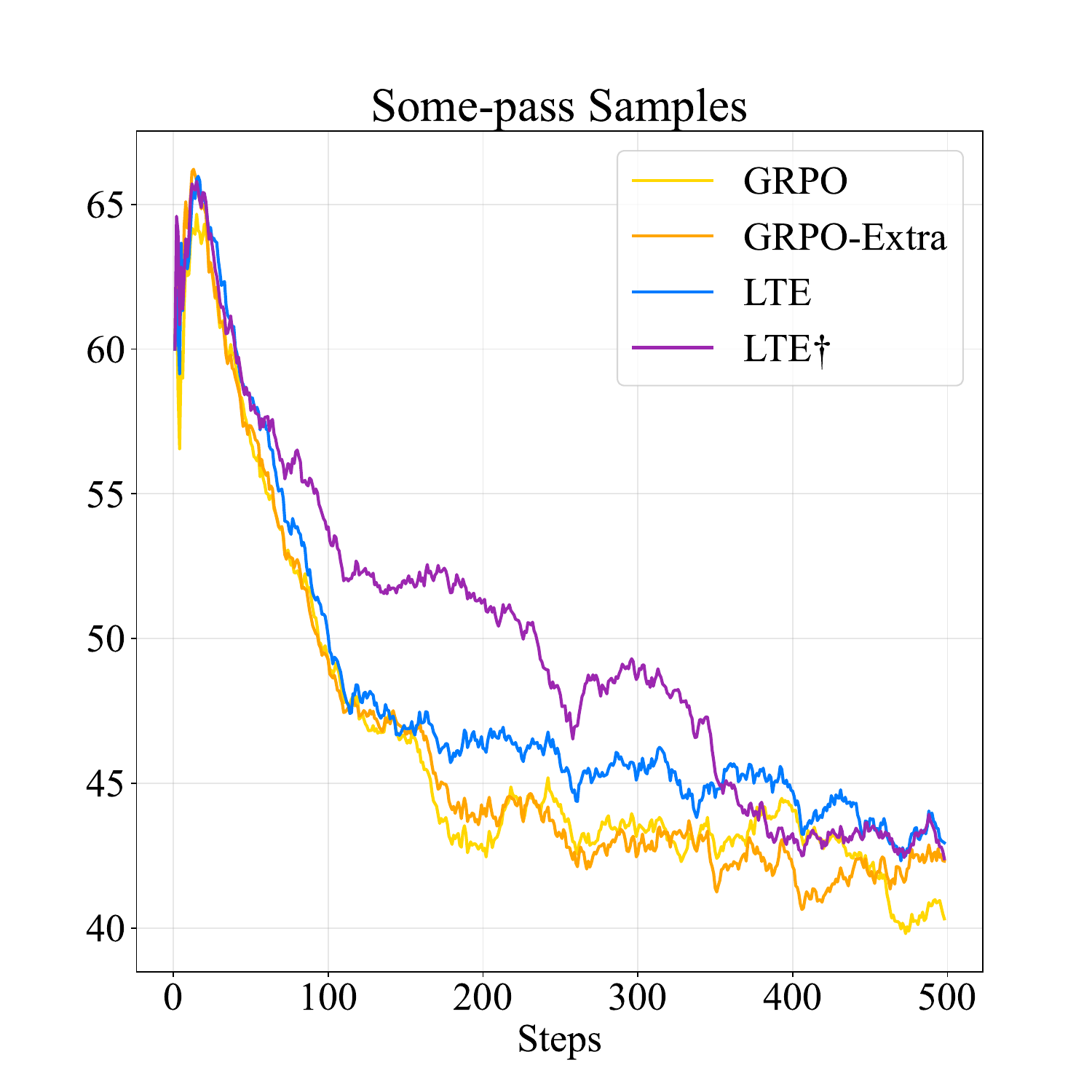}
        \caption{The number of \textit{some-pass} samples in the initial rollouts.}
        \label{subfig:some-pass}
    \end{subfigure}
    
    \caption{Performance of Qwen3-4B-Base on the training data, smoothed with exponential moving average.}
    \label{fig:training-data}
\end{figure}

We analyze the number of \textit{none-pass} and \textit{some-pass} samples of the training data to study whether and how exploration stagnation is addressed.

In Figure~\ref{subfig:none-pass}, the normal GRPO becomes stuck in exploration stagnation after about 150 training steps, in which the policy model cannot further solve a larger number of \textit{none-pass} samples any more. GRPO$_\text{Extra}$ also fails to break the spell. On the contrary, LTE and LTE$\dag$ consistently keep the LM solving more \textit{none-pass} samples during training, significantly reducing the number of unsolved problems even in the second half of training. This directly proves that LTE is successful in addressing the exploration stagnation issue in RLVR without any explicit external expert guidance.

Figure~\ref{subfig:some-pass} presents the number of \textit{some-pass} samples. During training, LTE and LTE$\dag$ keep a relatively higher level of number of \textit{some-pass} samples, which are non-zero-gradient groups that primarily bring learning signal. This indicates that LTE keeps a higher level of learnability~\cite{odf} during training, with more moderately uncertain \textit{some-pass} samples benefiting the learning.

The above observations prove that, only with hints from the LM's own experience, LTE successfully mitigates the issue of exploration stagnation and keeps a high learnability level during training.

\subsection{Analysis on the Training Dynamics}
\begin{figure*}[htbp]
\centering
    \begin{subfigure}[t]{0.245\textwidth}
        \centering
        \includegraphics[width=\linewidth]{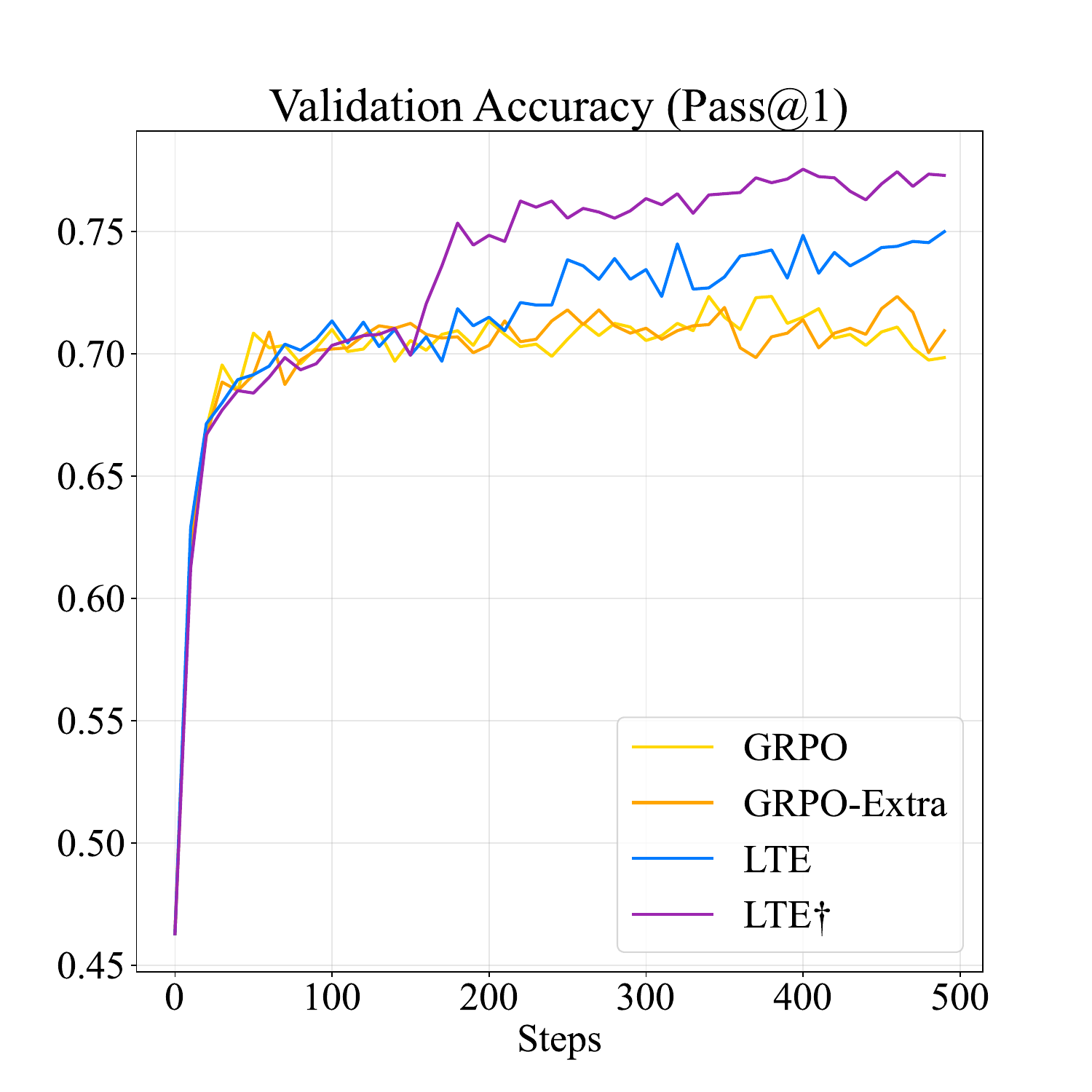}
        \caption{MATH-500 Pass@1.}
        \label{subfig:val-mean}
    \end{subfigure}
    \begin{subfigure}[t]{0.245\textwidth}
        \centering
        \includegraphics[width=\linewidth]{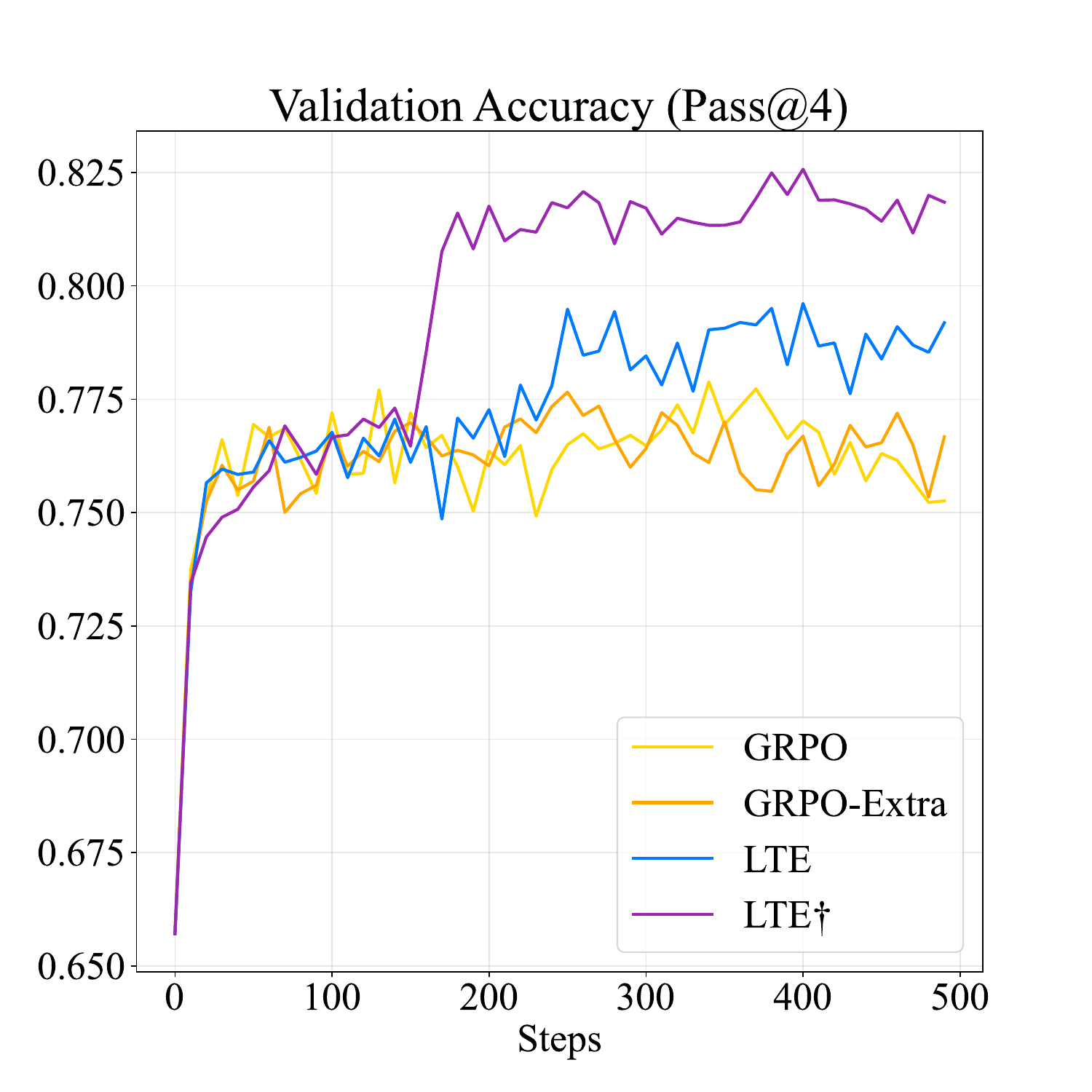}
        \caption{MATH-500 Pass@4.}
        \label{subfig:val-pass}
    \end{subfigure}
    \begin{subfigure}[t]{0.245\textwidth}
        \centering
        \includegraphics[width=\linewidth]{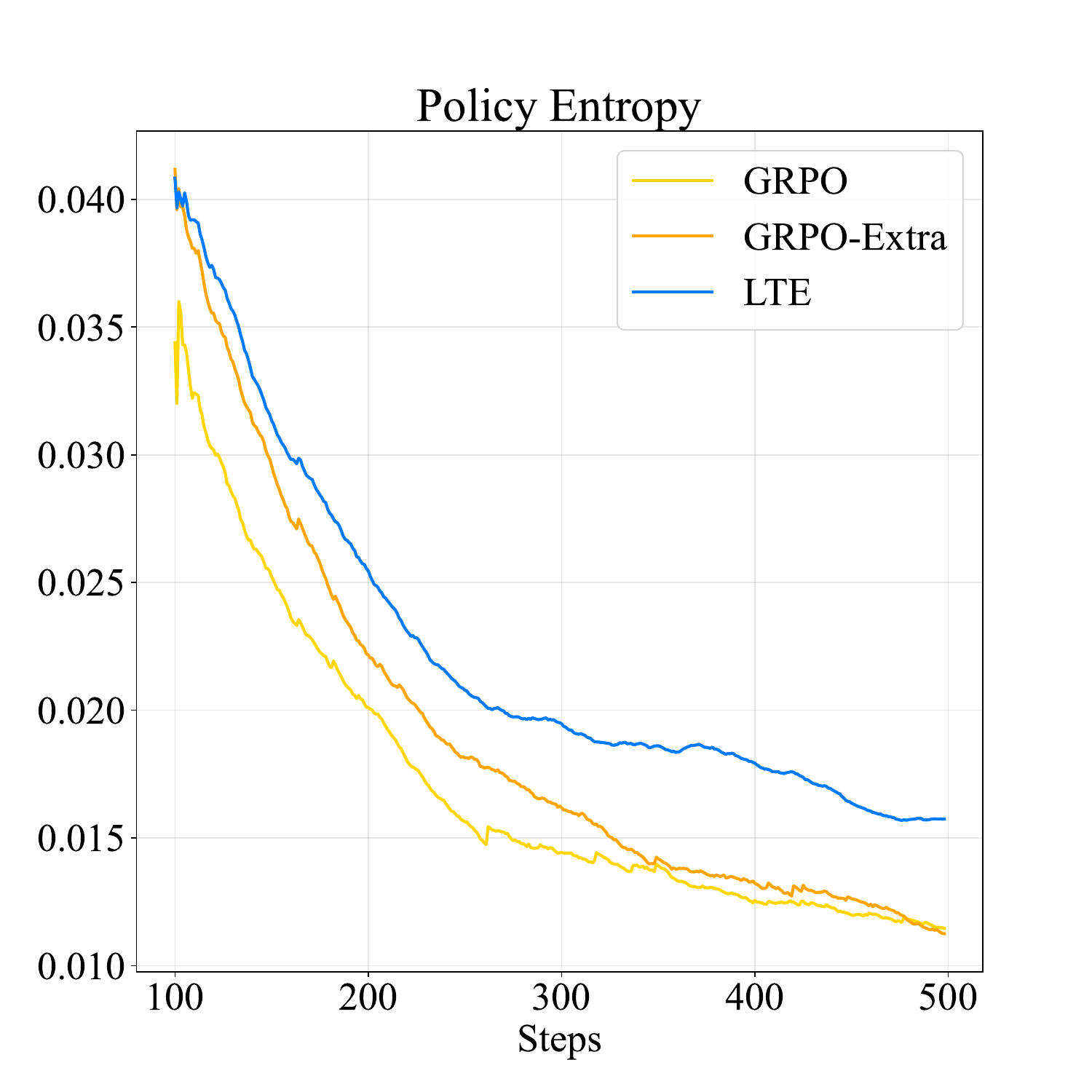}
        \caption{Policy entropy.}
        \label{subfig:entropy}
    \end{subfigure}
    \begin{subfigure}[t]{0.245\textwidth}
        \centering
        \includegraphics[width=\linewidth]{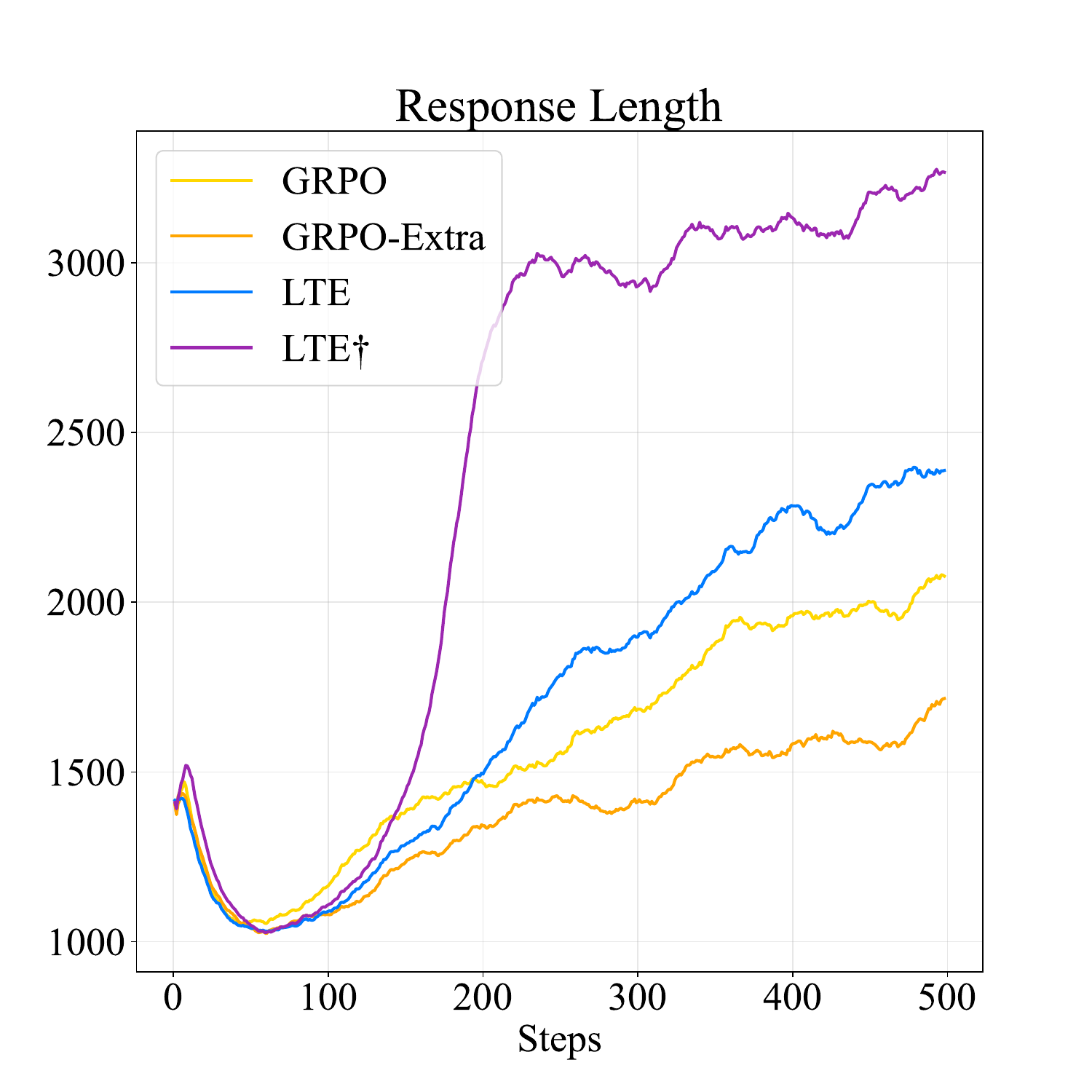}
        \caption{Response length.}
        \label{subfig:length}
    \end{subfigure}
    \caption{Qwen3-4B-Base training dynamics, with entropy and length smoothed by exponential moving average.}
    \label{fig:exploration}
\end{figure*}

In Figure~\ref{subfig:val-mean}, the Pass@1 on the validation data of GRPO and GRPO$_\text{Extra}$ get stagnated after 100 training steps with no further improvement. LTE and LTE$\dag$, however, continue to increase the validation score and converge to a higher level. In Figure~\ref{subfig:val-pass}, the Pass@4 scores of the two baselines even slightly shrink after 300 steps as the policy becomes over-certain. On the contrary, LTE and LTE$\dag$ keep unlocking the upper bound and achieving higher Pass@4. This further validates LTE's advantage in both exploitation and exploration.

As shown in Figure~\ref{subfig:entropy}, although all the three methods suffer from a trend of entropy collapse due to lack of explicit entropy control, the entropy of LTE remains relatively high throughout the long tail after 100 steps, indicating that it helps the LM keep a certain level of uncertainty and exploration.

From the perspective of response length presented in Figure~\ref{subfig:length}, we observe a noticeable trend. While GRPO and GRPO$_\text{Extra}$ exhibit a slow increase, LTE and LTE$\dag$, especially the latter, raise the LM's response length more significantly. This demonstrates that LTE implicitly encourages test-time deep thinking~\cite{r1} and drives the LM to spend more tokens on exploration, thus shaping a much more explorative policy.

The above analysis confirms that LTE effectively elicits the internal exploration capability in LMs while keeping an excellent level of exploitation.

\subsection{Cost Analysis}
\begin{table}[htbp]
\small
  \centering
    \begin{tabular}{lcc}
    \toprule
    \textbf{Method} & \textbf{EC} & \textbf{Time} \\
    \midrule
    GRPO  & -     & 5.0/7.0 \\
    GRPO$_\text{Extra}$ & -     & 5.5/7.0 \\
    EvoCoT & -     & 1.5/2.0 \\
    LTE   & -     & 6.0/9.0 \\
    \midrule
    ReLIFT & EL    & 9.0/12.5 \\
    DAPO  & CH    & 6.5/7.0 \\
    LUFFY & EL    & 5.5/8.0 \\
    LTE$\dag$  & EL    & 7.5/9.0 \\
    \bottomrule
    \end{tabular}%
  \caption{The training time (Qwen3-4B-Base/Qwen3-8B-Base) of all the methods in days.}
  \label{tab:cost}%
\end{table}%

The training time of all the evaluated methods is presented in Table~\ref{tab:cost}. Note that EvoCoT learns to output extremely short responses, leading to its extremly low time cost. However, as shown in Table~\ref{tab:pass-1} and \ref{tab:pass-k}, the learned policy of EvoCoT even cannot outperform the GRPO baseline. As shown, LTE requires two more days (around 30\%) than GRPO on average, while baselines like DAPO and LUFFY also requires about one extra day on average compared to GRPO. Overall, we believe such a moderate increase of cost is acceptable in practice (especially in industrial application), since we aim to breakthrough the upperbound of the LMs and the increase is not highly significant.

\begin{table}[htbp]
\small
  \centering
  \setlength{\tabcolsep}{1.0mm}
    \begin{tabular}{lccccc}
    \toprule
    \textbf{Setting} & \textbf{EC} & \textbf{Step} & \textbf{Time} & \textbf{Pass@1} & \textbf{Pass@k} \\
    \midrule
    \multicolumn{1}{l}{\textbf{\textsc{Qwen3-4B-Base}}} & -     & -     & -     & 22.59 & 48.31 \\
    \midrule
    GRPO  & -     & 500   & 7.0   & 34.76 & 50.45 \\
    GRPO$_\text{Extra}$ & -     & 500   & 7.0   & 36.47 & 50.70 \\
    LTE   & -     & 400   & 7.0   & \textbf{40.06} & \textbf{57.95} \\
    \midrule
    DAPO  & CH    & 500   & 7.0   & \textbf{44.45} & 59.74 \\
    LUFFY & EL    & 500   & 8.0   & 43.08 & 60.12 \\
    LTE$\dag$  & EL    & 300   & 7.0   & 43.71 & \textbf{62.83} \\
    \midrule
    \midrule
    \multicolumn{1}{l}{\textbf{\textsc{Qwen3-8B-Base}}} & -     & -     & -     & 30.39 & 52.87 \\
    \midrule
    GRPO  & -     & 500   & 5.0   & 43.99 & 56.82 \\
    GRPO$_\text{Extra}$ & -     & 500   & 5.5   & 41.72 & 56.74 \\
    LTE   & -     & 400   & 5.0   & \textbf{48.62} & \textbf{65.65} \\
    \midrule
    DAPO  & CH    & 500   & 6.5   & 45.40 & 60.58 \\
    LUFFY & EL    & 500   & 5.5   & 47.15 & \textbf{65.43} \\
    LTE$\dag$  & EL    & 400   & 5.0   & \textbf{47.75} & 63.36 \\
    \bottomrule
    \end{tabular}%
  \caption{Results with matching training cost averaged across all the six benchmarks. The time is in days.}
  \label{tab:analysis-cost}%
\end{table}%

% Since LTE requires extra rollouts for all \textit{none-pass} samples, it leads to extra overhead. As discussed in Appendix~\ref{app:cost}, LTE requires around 30\% more time than GRPO, which is acceptable since it aims to breakthrough the upperbound of the LMs.

We also perform a fair comparison with training cost almost aligned in Table~\ref{tab:cost}, in which we evaluate the intermediate steps of LTE to ensure it takes the least training time among all the methods. With 100 fewer training steps, LTE still significantly outperforms GRPO and GRPO$_\text{Extra}$. For instance, within both 5.0 days, Qwen3-8B-Base trained with LTE improves the Pass@1 and Pass@k of GRPO by \textbf{+4.63} and \textbf{+8.83}, respectively. When compared with advanced methods like DAPO and LUFFY, either with sophisticated technical tricks or external expert guidance, LTE$\dag$ exhibits competitive performance with 100 or 200 fewer steps. For instance, with Qwen3-4B-Base, LTE$\dag$ achieves higher Pass@1 and Pass@k than LUFFY (\textbf{+0.63} and \textbf{+2.71}, respectively) even with 1.0 fewer day. These confirm that, within the same budget, LTE helps the LMs learn more effectively; with extended training time, LTE continues to further enhance the LM.

\begin{table*}[htbp]
\setlength{\tabcolsep}{1.2mm}
\small
  \centering
    \begin{tabular}{lccccccc}
    \toprule
    \textbf{Setting} & \textbf{MATH-500} & \textbf{Minerva} & \textbf{Olympiad} & \textbf{AMC'23} & \textbf{AIME'24} & \textbf{AIME'25} & \textbf{Avg.} \\
    \midrule
    LTE   & 78.85/\textbf{85.20} & \textbf{37.87}/\textbf{47.06} & \textbf{45.52}/\textbf{54.22} & \textbf{73.91}/\textbf{97.50} & \textbf{30.62}/\textbf{70.00} & \textbf{27.29}/\textbf{46.67} & \textbf{49.01}/\textbf{66.78} \\
    \midrule
    \; w/o Incorrect & 76.95/83.20 & 36.67/44.85 & 41.59/51.85 & 69.53/95.00 & 25.00/60.00 & 17.08/40.00 & 44.47/62.48 \\
    \; w/o Concise & \textbf{79.75}/85.00 & 37.22/46.32 & 45.30/54.07 & 70.00/95.00 & 27.71/63.33 & 23.13/\textbf{46.67} & 47.19/65.07 \\
    \; w/o Either (GRPO$_\text{Extra}$) & 75.75/82.00 & 36.95/43.75 & 40.07/48.00 & 60.47/90.00 & 22.29/46.67 & 14.79/30.00 & 41.72/56.74 \\
    \bottomrule
    \end{tabular}%
  \caption{Ablation results (Pass@1/Pass@k) of Qwen3-8B-Base. Best results are \textbf{bolded}.}
  \label{tab:ablation}%
\end{table*}%

\subsection{Ablation Study}
We assess the indispensability of the types of hints in the prompts of LTE by ablating the source of hints (\textit{i.e.}, incorrect answers and concise hint). As shown in Table~\ref{tab:ablation}, both types of hints contribute to the performance of LTE, especially the hint of incorrect answers, which leads to a performance decline of \textbf{4.54} and \textbf{4.30} on average for Pass@1 and Pass@k when ablated. The concise hint is also essential with an averaged contribution of \textbf{1.82} and \textbf{1.71} for Pass@1 and Pass@k, respectively. Finally, when both types are ablated (\textit{i.e.}, GRPO$_\text{Extra}$), the trained policy suffers from a drastic degeneration (\textbf{7.29} for Pass@1 and \textbf{10.04} for Pass@k), confirming the hints are necessary for LTE.

Besides the different hints in prompts, we also validate the components regarding off-policy (Equation~\ref{eq:mixed-policy}) and importance sampling (Equation~\ref{eq:off-ratio}) of LTE's algorithm are necessary in Appendix~\ref{app:ablation-components}. Meanwhile, in Appendix~\ref{app:ps}, we demonstrate that the use of the policy-shaping trick (Equation~\ref{eq:shaping}) is not the source of LTE's improvement, indicating that LTE's outcoming performance comes from its core design instead of this technical trick.

\subsection{Effectiveness of LTE out of Mathematics}
\begin{table}[htbp]
\small
  \centering
    \begin{tabular}{lcc}
    \toprule
    \textbf{Method} & \textbf{Valid} & \textbf{Test} \\
    \midrule
    Base  & 15.50/28.89 & 17.60/32.37 \\
    GRPO  & 51.13/52.72 & 55.43/57.35 \\
    LTE   & \textbf{55.58}/\textbf{58.18} & \textbf{57.75}/\textbf{59.83} \\
    \bottomrule
    \end{tabular}%
  \caption{Results (Pass@1/Pass@4) of Qwen3-4B-Base in the science domain. "Valid" and "Test" refer to the validation and test set of SciQ, respectively.}
  \label{tab:sciq}%
\end{table}%

We also perform experiments on SciQ~\cite{sciq}, a dataset of science examination questions, to evaluate the performance of LTE in non-mathematical domains. All the settings follow our main experiments except for the maximum response length for evaluation, which is 8,192 for SciQ to be in line with the training.

As presented in Table~\ref{tab:sciq}, LTE outperforms GRPO in both Pass@1 and Pass@4 on both the validation set and the test set. These validates the effectiveness of LTE in the science domain. Nevertheless, the form of answers for this task is also short, allowing easy intergration with LTE. We leave applying LTE to other complex tasks that have longer outcomes or multi-turn interactions for future work.

\section{Related Work}\label{sec:related-work}
\subsection{Learning for Better Exploration}
Policy entropy is regarded as a metric indicating the exploration capability of LMs~\cite{entropy}. Some focus on preventing entropy collapse to enhance exploration. \citet{entropy} clip and apply KL penalty to high-covariance tokens. \citet{dapo} propose \texttt{Clip-Higher} to promote exploratory tokens. \citet{skywork} employ an adaptive entropy loss coefficient. Meanwhile, some others do not directly control the entropy. \citet{rrl} use historical replay to resume LMs' exploration. \citet{pass-at-k-training} propose Pass@k as the reward for training to balance exploration and exploitation. \citet{outcome-based-exploration} promote historical and batch diversity to encourage exploration.

\subsection{Learning from Guidance}
Some enhance RLVR using external reasoning traces which are either human-annotated or from stronger LMs by means of hybrid training~\cite{luffy, ReLIFT, chord, hpt} or hinting~\cite{StepHint, NuRL}, which suffer from limited accessibility and scalability as afore discussed. \citet{EvoCoT} propose EvoCoT, a guidance-based method that leverages only the ground truth label of training data as an explicit hint. LTE is different in that its hints consist of the LM's own trails instead of the ground truth, which fully exploits the LMs' previous experience and is more reward-hacking-free. \citet{laser} improve RLVR performance via the information from LMs' self-verification, which is another kind of self-guidance.

\subsection{Learning with More Rollouts}
\citet{brorl} perform extremely extended rollouts to break the exploration bottleneck.~\citet{knapsack} and~\citet{ar3po} both adopt an adaptive rollout strategy to assign more rollouts to difficult prompts, thus achieving training efficiency and exploration breakthrough at the same time. LTE is different from these in that it leverages the information from trial and error.

\subsection{Learning from Failures}
Some work also employs the idea of learning from failures but is essentially different from ours.~\citet{COPRA} include in-context execution feedback to perform multi-turn theorem-proving.~\citet{mistake-icl} include off-line mistakes generated by other LMs in the context to improve training-free reasoning.~\citet{multiattempt} performs iterative evaluation, verifying LMs' trial at test time and requiring to try again if it is incorrect, whose evaluation is indeed a variant of Pass@k. Similarly, \citet{ufo} focuses on multi-turn reasoning, in which the user says ``try again'' when the LM makes mistakes at test time. This setting assumes the user knows the ground truth which can be impractical in reality while LTE does not require any test-time verification.

\section{Conclusion and Future Work}\label{sec:conclusion}
We have presented LTE, a novel approach that aims to address the exploration stagnation of RLVR without explicit external expert guidance. It hints LMs with its previously generated incorrect answers to help better solve \textit{none-pass} samples. Experiments validate the effectiveness of LTE and further analysis confirms that LTE successfully mitigates exploration stagnation and elicits the internal exploration capability in LMs. There remain several avenues for future research including extending LTE to a broader set of more complex tasks, applying hints at more fine-grained intervals, and incorporating the correct ground truth into the hint while avoiding reward hacking.

% Future work includes:
% (1) designing an approach to incorporate the correct ground truth into the hint (while avoiding reward hacking) and make the hint more informative;
% (2) applying hints at more fine-grained intervals (\textit{e.g.}, each individual rollout receives a hint from accumulated incorrect answers).
% (3) extending LTE to a broader set of tasks besides mathematics.

\section*{Acknowledgments}
This work is supported by the National Natural
Science Foundation of China (62076008).

\section*{Limitations}
\paragraph{Domain} The current LTE instantiation only fits relatively short-horizon reasoning like mathematics and science due to the concise format of answers to these problems and requires further modification to support other complex tasks that have longer outcomes or require multi-turn interactions like coding and tool-use agents. One promising approach is to include a summarizer in the pipeline and employ the summaries of the previously failed interaction trajectories as the hints for the extra rollouts.

\paragraph{Training} Due to limited budget, we only train the LMs with a relatively short maximum response length (\textit{i.e.}, 8,192), which may not fully unlock the reasoning capability of the LMs.

\paragraph{Model} Due to limited computational resources, we have not experimented with larger-scale language models to study the performance of LTE on stronger policies.

\section*{Ethical Considerations}
\begin{table}[htbp]
\small
  \centering
    \begin{tabular}{ll}
    \toprule
    \textbf{Artifact} & \textbf{License} \\
    \midrule
    Qwen3 Models & Apache-2.0 \\
    verl  & Apache-2.0 \\
    Math-Verify & Apache-2.0 \\
    Elliott/Openr1-Math-46k-8192 & MIT \\
    MATH-500 & MIT \\
    Minerva & MIT \\
    OlympiadBench & MIT \\
    AMC'23 & N/A \\
    AIME'24 & N/A \\
    AIME'25 & N/A \\
    SciQ & CC BY-NC 3.0 \\
    EvoCoT & Apache-2.0 \\
    ReLIFT & N/A \\
    DAPO  & Apache-2.0 \\
    LUFFY & N/A \\
    \bottomrule
    \end{tabular}%
  \caption{Licenses of scientific artifacts we use.}
  \label{tab:license}%
\end{table}%

\paragraph{Use of AI Assistants}
The algorithmic design and core methodology of this work were derived through manual research and human reasoning. For coding, we work with the assistance of GitHub Copilot\footnote{\url{https://github.com/features/copilot}}. We certify that the substance of our code is our own work.

\paragraph{Computational Budget}
All our experiments are conducted on a machine with CentOS 8, 384 AMD$^\circledR$ EPYC\texttrademark{} 9K84 96-Core Processor CPUs and 2.2TiB memory. We use 8$\times$ NVIDIA H20 GPUs for all the experiments. The training of LTE/LTE$\dag$ takes around 6/7.5 and 9/9 days for the 4B and 8B models, respectively.

\paragraph{Reproducibility}
Our work is reproducible because we have provided our source code and implementation details.

\paragraph{Potential Risks}
To the best of our knowledge, there are no potential risks concerning our work.

\paragraph{Scientific Artifacts}
We cite all the creators of scientific artifacts we use in this paper. Licenses of these scientific artifacts are shown in Table~\ref{tab:license}. Our use of these artifacts is consistent with their intended use.


% \clearpage

\bibliography{custom}

\clearpage

\appendix

\section{Theoretical Analysis}
\label{app:theory}
\subsection{Definitions}
We follow the theoretical framework of CoT-Space~\cite{cot-space}, which reformulates the reasoning process of LMs as optimization in a continuous semantic manifold, to conduct theoretical analysis on LTE.

\begin{definition}[Reasoning State Space]
    Given a query $q$, $\mathcal{S}_q$ is the reasoning state space consisting of all possible intermediate states $s=(q, \xi)$, where $\xi=(\xi_1, \xi_2, ..., \xi_t)$ represents a partial reasoning chain with $t$ conceptual steps realized through multiple token generations.
\end{definition}

\begin{definition}[Minimums in Reasoning Space]
    Given a query $q$ and its grount truth answer $o^*$, assuming $\Xi_q$ is the set of all reasonable intermediate reasoning process for query $q$, the minimums of the reasoning space are defined as $\mathcal{M}_q=\{(q, \xi, o^*)\mid\xi\in\Xi_q\}$.
\end{definition}

% \begin{definition}[Reasoning Loss]
%     Given a query $q$ and its grount truth answer $o^*$, the loss function $C:\mathcal{S}_q\to\mathbb{R}^+$ quantifies the loss of each state $s$ towards the golden answer, which is positively correlated to the distance from $s$ to $o^*$. We have $C(m)=0$ for all all minimum states $m\in\mathcal{M}_q$, which are states with the correct answer $o^*$.
% \end{definition}

% For more details of reasoning state space and reasoning loss, please refer to ~\citet{cot-space}.

\begin{definition}[Failure and Pruned Subspace]
    Given a query $q$ with the incorrect answers $\mathcal{A}_q$ generated by the LM (for simplicity, we treat the answer of a truncated response as $\emptyset$), the failure subspace is $\mathcal{S}_q^{f}=\{s\in\mathcal{S}_q:\text{Extract}(s)\in\mathcal{A}_q\}$, where $\text{Extract}(\cdot)$ is an abstract function extracting the final answer from $s$. Then, the pruned subspace is $\mathcal{S}_q^{'} = \mathcal{S}_q\setminus\mathcal{S}_q^{f}$.
\end{definition}

\subsection{State Space Pruning}
In this section, we theoretically demonstrate that LTE achieves state space pruning with its hint of previous incorrect answers.

Starting from the law of total probability, we have
\begin{equation}\label{eq:total-prob-0}
\begin{aligned}
    &P(s \in \mathcal{M}_q) \\
    &= P(s \in \mathcal{M}_q | s \in \mathcal{S}_q^{'}) \cdot P(s \in \mathcal{S}_q^{'}) \\
    &\ \ \ \ + P(s \in \mathcal{M}_q | s \in \mathcal{S}_q^{f}) \cdot P(s \in \mathcal{S}_q^{f}).
\end{aligned}
\end{equation}
Since a correct answer cannot be in $\mathcal{S}_q^{f}$, we have
\begin{equation}
    P(s\in\mathcal{M}_q| s\in\mathcal{S}_q^{f}) = 0.
\end{equation}
Therefore, Equation~\ref{eq:total-prob-0} can be simplified to:
\begin{equation}\label{eq:total-prob-1}
    P(s \in \mathcal{M}_q) = P(s \in \mathcal{M}_q | s \in \mathcal{S}_q^{'}) \cdot P(s \in \mathcal{S}_q^{'})
\end{equation}

The probability of reaching a correct answer for $\pi_\theta$ given $q$ is:

\small
\begin{equation}
\begin{aligned}
    &P(s \in \mathcal{M}_q | q, \pi_\theta)\\
    &=P(s \in \mathcal{M}_q | s \in \mathcal{S}_q^{'}, q, \pi_\theta) \cdot P(s \in \mathcal{S}_q^{'}, q, \pi_\theta).
\end{aligned}
\end{equation}
\normalsize
When augmented with the hint information $\mathcal{H}_q$ containing $\mathcal{A}_q$ and the instruction of thinking concisely if $q$ is \textit{some-truncated} or \textit{all-truncated}, it becomes:

\small
\begin{equation}
\begin{aligned}
    &P(s \in \mathcal{M}_q | q, \mathcal{H}_q, \pi_\theta)\\
    &=P(s \in \mathcal{M}_q | s \in \mathcal{S}_q^{'}, q, \mathcal{H}_q, \pi_\theta) \cdot P(s \in \mathcal{S}_q^{'}, q, \mathcal{H}_q, \pi_\theta).
\end{aligned}
\end{equation}
\normalsize
The ratio of the above probabilities is:

\small
\begin{equation}\label{eq:prob-ratio-0}
\begin{aligned}
    &\frac{P(s \in \mathcal{M}_q | q, \mathcal{H}_q, \pi_\theta)}{P(s \in \mathcal{M}_q | q, \pi_\theta)}\\
    &=\frac{P(s \in \mathcal{M}_q | s \in \mathcal{S}_q^{'}, q, \mathcal{H}_q, \pi_\theta)}{P(s \in \mathcal{M}_q | s \in \mathcal{S}_q^{'}, q, \pi_\theta)} \cdot \frac{P(s \in \mathcal{S}_q^{'}, q, \mathcal{H}_q, \pi_\theta)}{P(s \in \mathcal{S}_q^{'}, q, \pi_\theta)}.
\end{aligned}
\end{equation}
\normalsize

Based on the instruction following capability of LMs, when given the hint $\mathcal{H}_q$, we may assume that the probability of the policy $\pi_\theta$ reaching a state $s$ in the failure subspace $\mathcal{S}_q^{f}$ decreases:
\begin{equation}
    P(s\in\mathcal{S}_q^{f}|q, \mathcal{H}_q, \pi_\theta) = P(s\in\mathcal{S}_q^{f}|q, \pi_\theta) - \delta,
\end{equation}
where $\delta>0$. Meanwhile, the probability of a space in the pruned subspace increases:
\begin{equation}
    P(s\in\mathcal{S}_q^{'}|q, \mathcal{H}_q, \pi_\theta) = P(s\in\mathcal{S}_q^{'}|q, \pi_\theta) + \delta.
\end{equation}

Let $\alpha$ denote $\frac{P(s \in \mathcal{M}_q | s \in \mathcal{S}_q^{'}, q, \mathcal{H}_q, \pi_\theta)}{P(s \in \mathcal{M}_q | s \in \mathcal{S}_q^{'}, q, \pi_\theta)}$, which indicates the ratio of the probability of reaching states in $\mathcal{M}_q$ when the policy reaches a state in $\mathcal{S}_q^{'}$ with or without $\mathcal{H}_q$. Then Equation~\ref{eq:prob-ratio-0} becomes:
\begin{equation}\label{eq:prob-ratio-1}
\begin{aligned}
    &\frac{P(s \in \mathcal{M}_q | q, \mathcal{H}_q, \pi_\theta)}{P(s \in \mathcal{M}_q | q, \pi_\theta)}\\
    &=\alpha \cdot \frac{P(s \in \mathcal{S}_q^{'}, q, \pi_\theta) + \delta}{P(s \in \mathcal{S}_q^{'}, q, \pi_\theta)}\\
    &=\alpha\cdot\left(1 + \frac{\delta}{P(s \in \mathcal{S}_q^{'}, q, \pi_\theta)}\right).
\end{aligned}
\end{equation}

For a \textit{none-pass} query $q$, all the $n$ initial rollouts lies in the failure subspace $\mathcal{S}_q^{f}$ according to the definitions. This allows us set a confidence level $\tau$ (where $0<\tau<1$) indicating the probability of observing $n$ consecutive failures:
\begin{equation}
    P(s \in \mathcal{S}_q^{f}, q, \pi_\theta)^n\ge\tau.
\end{equation}
According to the definitions of $\mathcal{S}_q^{f}$ and $\mathcal{S}_q^{'}$, we have $P(s \in \mathcal{S}_q^{f}, q, \pi_\theta) + P(s \in \mathcal{S}_q^{'}, q, \pi_\theta) = 1$. Thus
\begin{equation}
    P(s \in \mathcal{S}_q^{'}, q, \pi_\theta)\le1-\tau^{1/n}.
\end{equation}
Combining with Equation~\ref{eq:prob-ratio-1}, we have
\begin{equation}\label{eq:prob-ratio-2}
\begin{aligned}
    \frac{P(s \in \mathcal{M}_q | q, \mathcal{H}_q, \pi_\theta)}{P(s \in \mathcal{M}_q | q, \pi_\theta)}\ge\alpha\cdot\left(1 + \frac{\delta}{1-\tau^{1/n}}\right).
\end{aligned}
\end{equation}

Since the hint $\mathcal{H}_q$ should have little negative effect on the policy's reasoning capability in the pruned state space $\mathcal{S}^{'}_q$, we can reasonably assume that the possibility of $s\in\mathcal{M}_q$ when $s\in\mathcal{S}^{'}_q$ does not significantly decrease when it is conditioned on $\mathcal{H}_q$. This indicates that $\alpha$ should not be much less than $1$ (\textit{i.e.}, we may at least assume $\alpha>\Omega(\frac{1}{n})$). According to Equation~\ref{eq:prob-ratio-2}, if $\frac{P(s \in \mathcal{M}_q | q, \mathcal{H}_q, \pi_\theta)}{P(s \in \mathcal{M}_q | q, \pi_\theta)} \le 1$, then $\alpha \le 1/(1+\frac{\delta}{1-\tau^{1/n}})$. Using the approximation $\tau^{1/n} = e^\frac{\ln\tau}{n}\approx1+\frac{\ln\tau}{n}$, we approximately have $\alpha \le 1/(1+\frac{\delta}{|\ln\tau|}\cdot n)\sim\Theta(\frac{1}{n})$, which contradicts to our assumption. Therefore, we have
\begin{equation}\label{eq:prob-ratio-3}
\begin{aligned}
    \frac{P(s \in \mathcal{M}_q | q, \mathcal{H}_q, \pi_\theta)}{P(s \in \mathcal{M}_q | q, \pi_\theta)} > 1.
\end{aligned}
\end{equation}

This theoretically demonstrates that with the help of hint based on self-generated incorrect answers, LTE increases the possibility of reaching the correct answer in the extra rollouts, thus increasing the chance of the policy receiving meaningful training signals from previously \textit{none-pass} samples.

\section{Necessity of Random Replacement}\label{app:random-replacement}

The random replacement of rollouts in Section~\ref{subsec:mix-policy} is a natural and necessary design for the extra rollouts instead of a simple technical trick because there are no other reasonable ways of fusing the extra rollouts into the training.

If we simply insert the extra rollouts into the initial group without random replacement, the size of the group will be inconsistent across different training samples. For instance, if our initial group contains 8 rollouts, then the final groups will have 8-16 rollouts depending on the number of correct extra rollouts. This inconsistency will make the training unstable. Moreover, in engineering practice, the inconsistent group size will also lead to a severe technical issue related to the implementation of distributed training, which typically requires a consistent group size that can be dividible by the number of training backend workers (usually the number of GPUs, \textit{e.g.}, 8 in our case). A forced break of this consistency will cause a lot of engineering efforts and make the training less efficient. Therefore, we believe the random replacement of rollouts is a necessary design of our method instead of an easily dispensable trick.

\section{Implementation Details}
\label{app:detail}

\begin{table}[htbp]
  \centering
    \begin{tabular}{lc}
    \toprule
    \textbf{Hyper-parameter} & \textbf{Value} \\
    \midrule
    Learning Rate & 1e-6 \\
    Totel Steps & 500 \\
    Batch Size & 128 \\
    Mini Batch Size & 32 \\
    Micro Batch Size & 1.0 \\
    KL Loss Coefficient & 0.001 \\
    Clip Ratio & 0.2 \\
    Temperature & 1 \\
    Number of Rollouts & 8 \\
    Maximum Prompt Length & 2048 \\
    Maximum Response Length & 8192 \\
    \bottomrule
    \end{tabular}
  \caption{Full hyper-parameters for training.}
  \label{tab:hparam}
\end{table}

Following~\citet{luffy}, in our experiments, we set $\gamma=0.1$ for Equation~\ref{eq:shaping}. For computational efficiency and simplicity, we treat the old policy term $\pi_{\theta_{\mathrm{old}}}\bigl(o'_{i,t}\mid \mathcal{H}_q, q,\, o_{i,<t}\bigr)$ in the importance sampling ratio as $1$.

The full hyper-parameters used in our training are listed in Table~\ref{tab:hparam}. We set the coefficient of the entropy loss as 0.003 for both LUFFY and LTE$\dag$. For all the methods, we evaluate their final checkpoints after 500 steps.

For both training and evaluation, we verify the correctness of answers with \texttt{Math-Verify}\footnote{\url{https://github.com/huggingface/Math-Verify}}.

% \section{Training Cost}
% \label{app:cost}
% \input{tables/cost}

% The training time of all the evaluated methods is presented in Table~\ref{tab:cost}. Note that EvoCoT learns to output extremely short responses, leading to its extremly low time cost. However, as shown in Table~\ref{tab:pass-1} and \ref{tab:pass-k}, the learned policy of EvoCoT even cannot outperform the GRPO baseline.

% As shown, LTE requires two more days (around 30\%) than GRPO on average, while baselines like DAPO and LUFFY also requires about one extra day on average compared to GRPO. Overall, we believe such a moderate increase of cost is acceptable in practice (especially in industrial application), since we aim to breakthrough the upperbound of the LMs and the increase is not highly significant.

\begin{table*}[htbp]
  \centering
  \setlength{\tabcolsep}{1.2mm}
  \small
    \begin{tabular}{lccccccc}
    \toprule
    \textbf{Setting} & \textbf{MATH-500} & \textbf{Minerva} & \textbf{Olympiad} & \textbf{AMC'23} & \textbf{AIME'24} & \textbf{AIME'25} & \textbf{Avg.} \\
    \midrule
    LTE   & \textbf{74.55}/\textbf{80.60} & \textbf{34.19}/\textbf{41.91} & \textbf{39.78}/\textbf{49.48} & 56.72/\textbf{85.00} & \textbf{18.33}/30.00 & 14.17/\textbf{46.67} & \textbf{39.62}/\textbf{55.61} \\
    \midrule
    \; w/o OP & 73.60/79.20 & 32.35/40.07 & 37.26/46.37 & \textbf{57.34}/\textbf{85.00} & 13.12/33.33 & \textbf{15.00}/40.00 & 38.11/54.00 \\
    \; w/o IS & 46.00/72.20 & 19.21/33.82 & 20.78/36.59 & 32.50/80.00 & 10.21/\textbf{36.67} & 3.33/23.33 & 22.01/47.10 \\
    \midrule
    Base  & 45.40/69.80 & 19.49/37.87 & 22.81/39.70 & 35.31/82.50 & 8.75/33.33 & 3.75/26.67 & 22.59/48.31 \\
    \bottomrule
    \end{tabular}%
  \caption{Results ablating LTE's off-policy (OP) and importance sampling (IS) components (Pass@1/Pass@k) of Qwen3-4B-Base. Best results are \textbf{bolded}.}
  \label{tab:ablation-optimization}%
\end{table*}%

\begin{table*}[htbp]
  \centering
  \setlength{\tabcolsep}{1.2mm}
  \small
    \begin{tabular}{lccccccc}
    \toprule
    \textbf{Setting} & \textbf{MATH-500} & \textbf{Minerva} & \textbf{Olympiad} & \textbf{AMC'23} & \textbf{AIME'24} & \textbf{AIME'25} & \textbf{Avg.} \\
    \midrule
    LTE   & 74.55/80.60 & \textbf{34.19}/\textbf{41.91} & \textbf{39.78}/\textbf{49.48} & 56.72/85.00 & \textbf{18.33}/30.00 & 14.17/\textbf{46.67} & 39.62/55.61 \\
    LTE w/o PS & \textbf{75.10}/\textbf{82.00} & 32.54/41.54 & 38.56/48.15 & \textbf{60.16}/\textbf{95.00} & 17.29/\textbf{40.00} & \textbf{17.92}/36.67 & \textbf{40.26}/\textbf{57.23} \\
    \bottomrule
    \end{tabular}%
  \caption{Results (Pass@1/Pass@k) showing the effect of the policy shaping (PS) trick with Qwen3-4B-Base.}
  \label{tab:analysis-ps}%
\end{table*}%

\section{Necessity of Algorithm Components}
\label{app:ablation-components}
We also demonstrate the necessity of the design of off-policy (OP) optimization (for hinted trajectories) and importance sampling (IS) in the algorithm of LTE. For the ablation of OP, we treat the hinted trajectories as on-policy ones and directly use GRPO (Equation~\ref{eq:grpo}) instead of the mixed-policy version (Equation~\ref{eq:mixed-policy}). For the ablation of IS, we set the value of the IS ratio (Equation~\ref{eq:ratio}) to 1. As shown in Table~\ref{tab:ablation-optimization}, if the hinted trajectories are not treated in an OP manner, the overall performance declines (\textbf{1.51} for Pass@1 and \textbf{1.61} for Pass@k). Meanwhile, if we do not apply IS, the training almost fails and the LM scarcely learns anything, with the final performance even worse than the base model. These confirm that the OP and IS components in LTE's algorithm are necessary.

\section{Effect of Policy Shaping}
\label{app:ps}
By default, LTE applies the policy shaping (PS) trick (Equation~\ref{eq:shaping}) following~\citet{luffy}. As shown in Table~\ref{tab:analysis-ps}, whether trained with PS or not, LTE consistently exhibits outstanding performance. The averaged scores of LTE without PS are even higher that LTE with PS. This demonstrates that the technical trick does not contribute to the performance of LTE. In other words, the performance improvement is from the core design of LTE instead of this technical trick.

\end{document}